\documentclass[10pt,twocolumn,letterpaper]{article}

\usepackage{cvpr}
\usepackage{times}
\usepackage{epsfig}
\usepackage{graphicx}
\usepackage{amsmath}
\usepackage{amssymb}

\usepackage{multirow}
\usepackage{booktabs}
\usepackage{subcaption}

\usepackage[pagebackref=true,breaklinks=true,letterpaper=true,colorlinks,bookmarks=false]{hyperref}

\cvprfinalcopy 

\def\cvprPaperID{8361} 

\ifcvprfinal\fi
\begin{document}

\title{Learning Disentangled Representations via Mutual Information Estimation}

\author{\textbf{Eduardo H. Sanchez}\\
IRT Saint Exup\'{e}ry, IRIT\\
Toulouse, France\\
\and
\textbf{Mathieu Serrurier}\\
IRIT\\
Toulouse, France\\
\and
\textbf{Mathias Ortner}\\
IRT Saint Exup\'{e}ry\\
Toulouse, France\\
}

\maketitle

\begin{abstract}
In this paper, we investigate the problem of learning disentangled representations. Given a pair of images sharing some attributes, we aim to create a low-dimensional representation which is split into two parts: a shared representation that captures the common information between the images and an exclusive representation that contains the specific information of each image. To address this issue, we propose a model based on mutual information estimation without relying on image reconstruction or image generation. Mutual information maximization is performed to capture the attributes of data in the shared and exclusive representations while we minimize the mutual information between the shared and exclusive representation to enforce representation disentanglement. We show that these representations are useful to perform downstream tasks such as image classification and image retrieval based on the shared or exclusive component. Moreover, classification results show that our model outperforms the state-of-the-art model based on VAE/GAN approaches in representation disentanglement. 
\end{abstract}

\section{Introduction}\label{sec:Intro}

Deep learning success involves supervised learning where massive amounts of labeled data are used to learn useful representations from raw data. As labeled data is not always accessible, unsupervised learning algorithms have been proposed to learn useful data representations easily transferable for downstream tasks. A desirable property of these algorithms is to perform dimensionality reduction while keeping the most important attributes of data. For instance, methods based on deep neural networks have been proposed using autoencoder approaches \cite{higgins2017beta,pmlr-v80-kim18b,Kingma2013} or generative models \cite{arjovsky2017wasserstein,donahue2016adversarial,Goodfellow2014,pmlr-v48-larsen16,Mao2017,radford2015unsupervised}. Nevertheless, learning high-dimensional data can be challenging. Autoencoders present some difficulties to deal with multimodal data distributions and generative models rely on computationally demanding models \cite{gonzalez2018image,karras2018style,park2019SPADE} which are particularly complicated to train.

Recent work has focused on mutual information estimation and maximization to perform representation learning \cite{belghazi2018mine,hjelm2018learning,oord2018representation,ozair2019wasserstein}. As mutual information maximization is shown to be effective to capture the salient attributes of data, another desirable property is to be able to disentangle these attributes. For instance, it could be useful to remove some attributes of data that are not relevant for a given task such as illumination conditions in object recognition. 

In particular, we are interested in learning representations of data that shares some attributes. Learning a representation that separates the common data attributes from the remaining data attributes could be useful in multiple situations. For example, capturing the common information from multiple face images could be advantageous to perform pose-invariant face recognition \cite{tran2017disentangled}. Similarly, learning a representation containing the common information across satellite image time series is shown to be useful for image classification and segmentation \cite{SanchezDisentanglingSatellite}. 

\begin{figure}[t]
\centering
\includegraphics[width=1.00\linewidth]{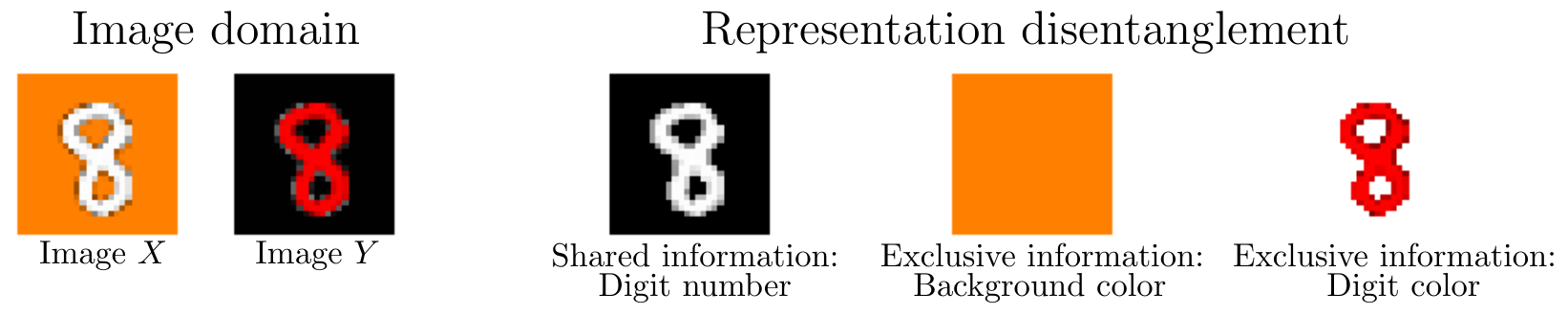}
\caption{Representation disentanglement example. Given images $X$ and $Y$ on the left, our model aims to learn a representation space where the image information is split into the shared information (digit number) and the exclusive information (background/digit color) on the right.}
\label{fig:RepresentationDisentanglementExample}
\end{figure}

In this paper, we propose a method to learn disentangled representations based on mutual information estimation. Given an image pair (typically from different domains), we aim to disentangle the representation of these images into two parts: a shared representation that captures the common information between images and an exclusive representation that contains the specific information of each image. An example is shown in Figure \ref{fig:RepresentationDisentanglementExample}. To capture the common information, we propose a novel method called \textit{cross mutual information estimation and maximization}. Additionally, we propose an adversarial objective to minimize the mutual information between the shared and exclusive representations in order to achieve representation disentanglement. The following contributions are made in this work:

\begin{itemize}
\item Based on mutual information estimation (see Section \ref{sec:Background}), we propose a method to learn disentangled representations without relying on more costly image reconstruction or image generation models.
\item In Section \ref{sec:Method}, we present a novel training procedure which is divided into two stages. First, the shared representation is learned via \textit{cross mutual information estimation and maximization}. Secondly, mutual information maximization is performed to learn the exclusive representation while minimizing the mutual information between the shared and exclusive representations. We introduce an adversarial objective to minimize the mutual information as the method based on statistics networks described in Section \ref{sec:Background} is not suitable for this purpose.
\item In Section \ref{sec:Experiments}, we perform several experiments on two synthetic datasets: a) colored-MNIST \cite{lecun-mnisthandwrittendigit-2010}; b) 3D Shapes \cite{3dshapes18} and two real dataset: c) IAM Handwriting \cite{marti2002iam}; d) Sentinel-2 \cite{drusch2012sentinel}. We show that the obtained representations are useful at image classification and image retrieval outperforming the state-of-the-art model based on VAE/GAN approaches in representation disentanglement. We perform an ablation study to analyze some components of our model. We also show the effectiveness of the proposed adversarial objective in representation disentanglement via a sensitivity analysis. In Section \ref{sec:Conclusions}, we show the conclusions of our work.
\end{itemize}

\section{Related work}\label{sec:RelatedWork}

\paragraph{Generative adversarial networks (GANs)}
The GAN model \cite{Goodfellow2014,Goodfellow2016} can be thought of as an adversarial game between two players: the generator and the discriminator. In this setting, the generator aims to produce samples that look like drawn from the data distribution $\mathbb{P}_{data}$ while the discriminator receives samples from the generator and the dataset to determine their source (dataset samples from $\mathbb{P}_{data}$ or generated samples from $\mathbb{P}_{gen}$). The generator is trained to fool the discriminator by learning a distribution $\mathbb{P}_{gen}$ that converges to $\mathbb{P}_{data}$. 

\paragraph{Mutual information}
Recent work has focused on mutual information estimation and maximization as a means to perform representation learning. Since the mutual information is notoriously hard to compute for high-dimensional variables, some estimators based on deep neural networks have been proposed. Belghazi \etal \cite{belghazi2018mine} propose a mutual information estimator which is based on the Donsker-Varadhan representation of the Kullback-Leibler divergence. Instead, Hjelm \etal \cite{hjelm2018learning} propose an objective function based on the Jensen-Shannon divergence called Deep InfoMax. Similarly, Ozair \etal \cite{ozair2019wasserstein} use the Wasserstein divergence. Mutual information maximization based methods learn representations without training decoder functions that go back into the image domain which is the prevalent paradigm in representation learning.

\paragraph{Representation disentanglement}
Disentangling data attributes can be useful for several tasks that require knowledge of these attributes. Creating representations where each dimension is independent and corresponds to a particular attribute have been proposed using VAE variants \cite{higgins2017beta,pmlr-v80-kim18b}. Chen \etal \cite{Chen2016} propose a GAN model combined with a mutual information regularization. Similar to our work, Gonzalez-Garcia \etal \cite{gonzalez2018image} propose a model based on VAE-GAN image translators and gradient reversal layers \cite{ganin2014unsupervised} to disentangle the attributes of paired data into shared and exclusive representations. 

\paragraph{}
In this work, we aim to learn disentangled representations of paired data by splitting the representation into a shared part and an exclusive part. We propose a model based on mutual information estimation to perform representation learning using the method of Hjelm \etal \cite{hjelm2018learning} instead of generative or autoencoding models. Additionally, we introduce an adversarial objective \cite{Goodfellow2014} to disentangle the information contained in the shared and exclusive representations which is more effective than the gradient reversal layers \cite{ganin2014unsupervised}. We compare our model to the model proposed by Gonzalez-Garcia \etal \cite{gonzalez2018image} since we have a common goal: to disentangle the representation space into a shared and an exclusive representation for paired data. We show that we achieve better results for representation disentanglement.

\section{Mutual information}\label{sec:Background}

Let $X \in \mathcal{X}$ and $Z \in \mathcal{Z}$ be two random variables. Assuming that $p(x,z)$ is the joint probability density function of $X$ and $Z$ and that $p(x)$ and $p(z)$ are the corresponding marginal probability density functions, the mutual information between $X$ and $Z$ can be expressed as follows
\begin{equation}
I(X, Z) = \int_{\mathcal{X}} \int_{\mathcal{Z}} p(x,z) \log \left( \frac{p(x,z)}{p(x)p(z)} \right) dxdz
\label{eq:MutualInformationDefinition}
\end{equation}
From Equation \ref{eq:MutualInformationDefinition}, it can be seen that the mutual information $I(X, Z)$ can be written as the Kullback-Leibler divergence between the joint probability distribution $\mathbb{P}_{XZ}$ and the product of the marginal distributions $\mathbb{P}_{X}\mathbb{P}_{Z}$, \ie $I(X, Z) = \text{D}_{KL} \left(\mathbb{P}_{XZ} \parallel \mathbb{P}_{X}\mathbb{P}_{Z} \right)$. 

In this work, we use the mutual information estimator Deep InfoMax \cite{hjelm2018learning} where the objective function is based on the Jensen-Shannon divergence instead, \ie $I^{\text{(JSD)}}(X, Z) = \text{D}_{JS} \left(\mathbb{P}_{XZ} \parallel \mathbb{P}_{X}\mathbb{P}_{Z} \right)$. We employ this method since it proves to be stable and we are not interested in the precise value of mutual information but in maximizing it. The estimator is shown in Equation \ref{eq:MutualInformationJSEstimator} where $T_{\theta}: \mathcal{X} \times \mathcal{Z} \rightarrow \mathbb{R}$ is a deep neural network of parameters ${\theta}$ called the \textit{statistics network}.
\begin{equation}
\begin{split}
\hat{I}_{\theta}^{\text{(JSD)}}(X, Z) &= \mathbb{E}_{p(x,z)} \left[ -\log \left( 1 + e^{- T_{\theta}(x,z)} \right) \right] \\
& \quad - \mathbb{E}_{p(x)p(z)} \left[ \log \left( 1 + e^{T_{\theta}(x,z)} \right) \right]
\end{split}
\label{eq:MutualInformationJSEstimator}
\end{equation}
Hjelm \etal \cite{hjelm2018learning} propose an objective function based on the estimation and maximization of the mutual information between an image $X \in \mathcal{X}$ and its feature representation $Z \in \mathcal{Z}$ which is called \textit{global mutual information}. The feature representation $Z$ is extracted by a deep neural network of parameters $\psi$, $E_{\psi}:\mathcal{X} \rightarrow \mathcal{Z}$. Equation \ref{eq:MutualInformationJensenShannonGlobal} displays the global mutual information objective.
\begin{equation}
\mathbf{L}^{\mathrm{global}}_{\theta, \psi}(X,Z) = \hat{I}_{\theta}^{\text{(JSD)}}(X, Z)
\label{eq:MutualInformationJensenShannonGlobal}
\end{equation}
Additionally,  Hjelm \etal \cite{hjelm2018learning} propose to maximize the mutual information between local patches of the image $X$ represented by a feature map $C_{\psi}(X)$ of the encoder $E_{\psi} = f_{\psi} \circ C_{\psi}$ and the feature representation $Z$ which is called \textit{local mutual information}. Equation \ref{eq:MutualInformationJensenShannonLocal} shows the local mutual information objective.
\begin{equation}
\mathbf{L}^{\mathrm{local}}_{\phi, \psi}(X,Z) = \hat{I}_{\phi}^{\text{(JSD)}}(C_{\psi}(X), Z)
\label{eq:MutualInformationJensenShannonLocal}
\end{equation}

\begin{figure*}[h!]
\centering
\begin{subfigure}[b]{0.38\textwidth}
\centering
\includegraphics[width=\textwidth]{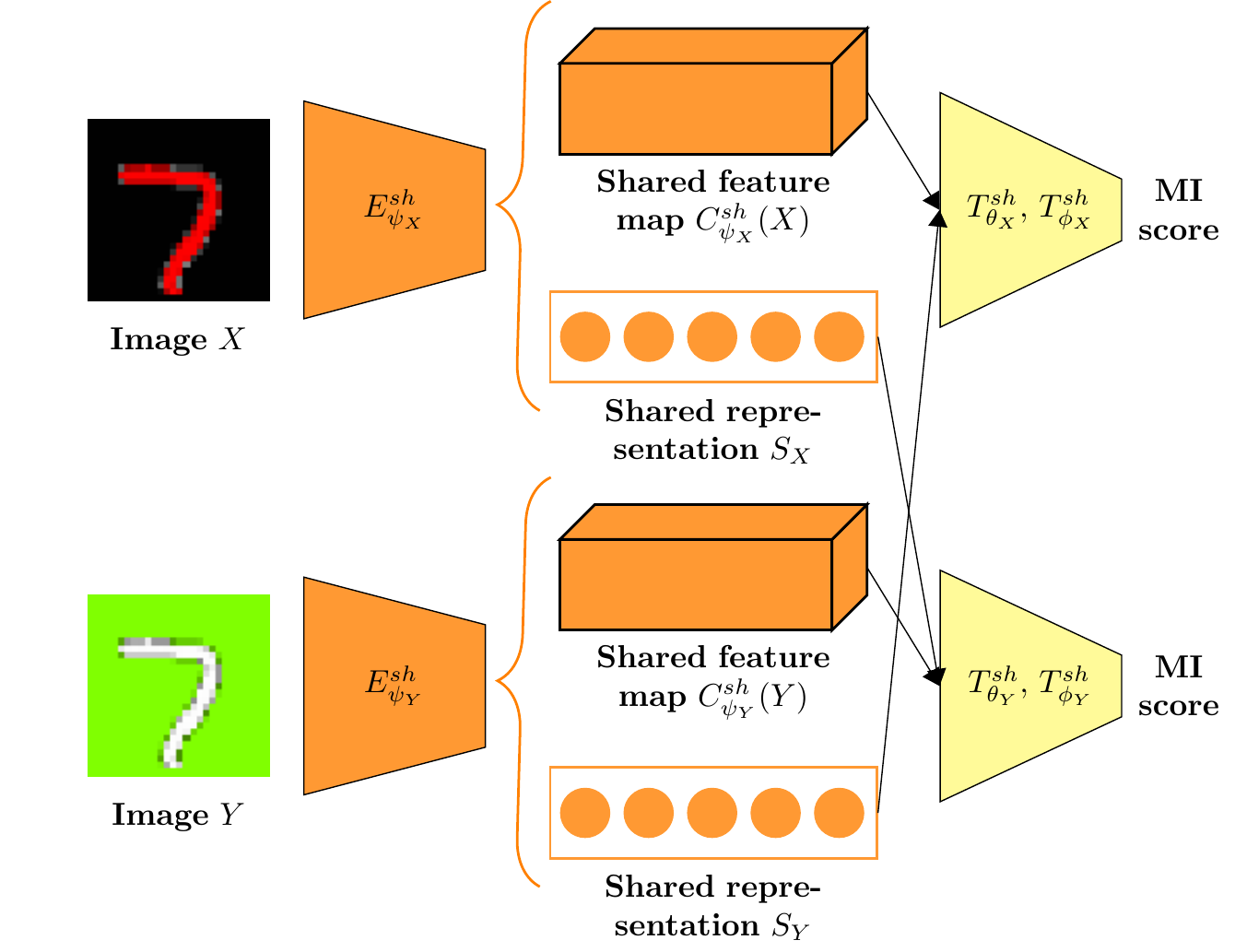}
\caption{}
\label{}
\end{subfigure}
\hfill
\begin{subfigure}[b]{0.60\textwidth}
\centering
\includegraphics[width=\textwidth]{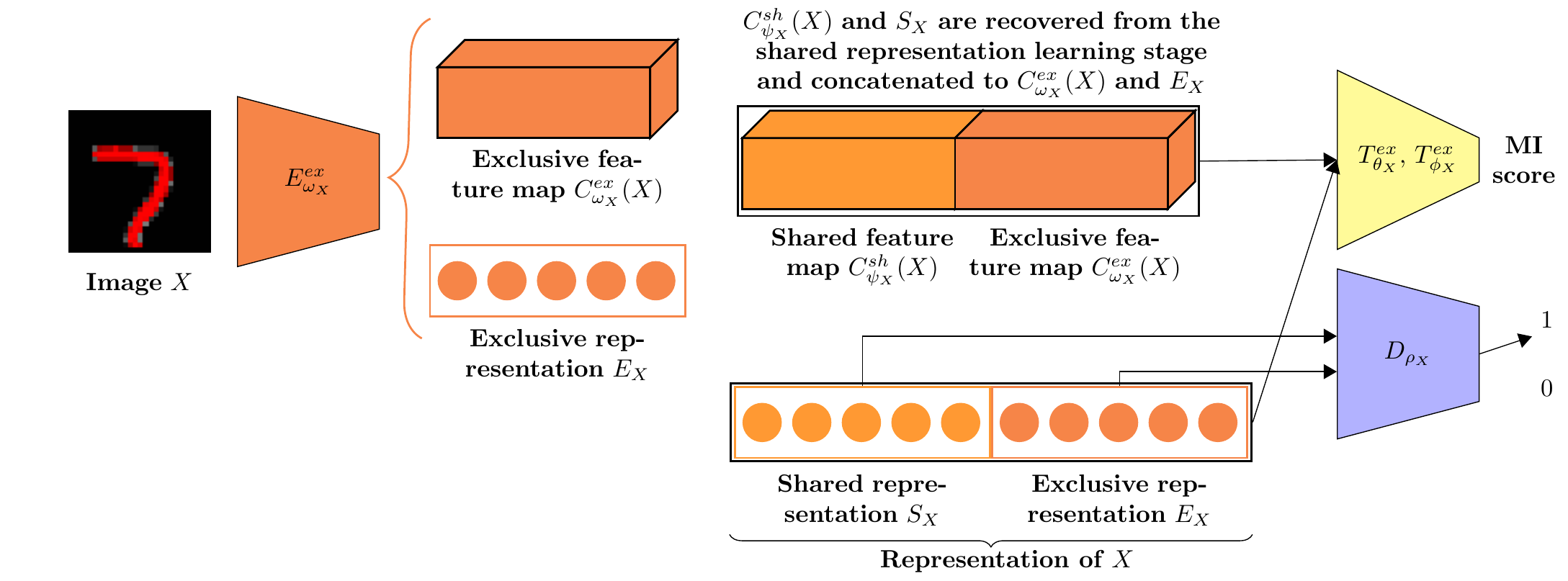}
\caption{}
\label{}
\end{subfigure}
\caption{Model overview. a) First, the shared representation is learned. Images $X$ and $Y$ are passed through the shared representation encoders to extract the representations $S_X$ and $S_Y$. The statistics networks maximize the mutual information between the image $X$ and the representation $S_Y$ (and between $Y$ and $S_X$); b) Then, the exclusive representation is learned. The image $X$ is passed through the exclusive representation encoder to obtain the representation $E_X$. The statistics networks maximize the mutual information between the image $X$ and its representation $R_X = (S_X, E_X)$ while the discriminator minimize the mutual information between representations $S_X$ and $E_X$. The same operation is performed to learn $E_Y$.}
\label{fig:Model}
\end{figure*}

\section{Method}\label{sec:Method}

Let $X$ and $Y$ be two images belonging to the domains $\mathcal{X}$ and $\mathcal{Y}$ respectively. Let $R_X \in \mathcal{R_X}$ and $R_Y \in \mathcal{R_Y}$ be the corresponding representations for each image. The representation is split into two parts: the shared representations $S_X$ and $S_Y$ which contain the common information between the images $X$ and $Y$ and the exclusive representations $E_X$ and $E_Y$ which contain the specific information of each image. Therefore the representation of image $X$ can be written as $R_X = (S_X, E_X)$. Similarly,  we can write $R_Y = (S_Y, E_Y)$ for image $Y$. For instance, let us consider the images shown in Figure \ref{fig:RepresentationDisentanglementExample}. In this case, the shared representations $S_X$ and $S_Y$ contain the digit number information while the exclusive representations $E_X$ and $E_Y$ correspond to the background and digit color information.

To address this representation disentanglement, we propose a training procedure which is split into two stages. We think that a natural way to learn these disentangled representations can be done via an incremental approach. The first stage learns the common information between images and creates a shared representation (see Section \ref{subsec:SharedLearning}). Knowing the common information, it is easy then to identify the specific information of each image. Therefore, using the shared representation from the previous stage, a second stage is then performed to learn the exclusive representation (see Section \ref{subsec:ExclusiveLearning}) which captures the remaining information that is missing in the shared representation. The model overview can be seen in Figure \ref{fig:Model}.

\subsection{Shared representation learning}\label{subsec:SharedLearning}

Let $E_{\psi_X}^{\mathrm{sh}}:\mathcal{X} \rightarrow \mathcal{S_X}$ and $E_{\psi_Y}^{\mathrm{sh}}:\mathcal{Y} \rightarrow \mathcal{S_Y}$ be the encoder functions to extract the shared representations $S_X$ and $S_Y$ from images $X$ and $Y$, respectively. We estimate and maximize the mutual information between the images and their shared representations via Equations \ref{eq:MutualInformationJensenShannonGlobal} and \ref{eq:MutualInformationJensenShannonLocal} using the global statistics networks $T_{\theta_X}^{\mathrm{sh}}$ and $T_{\theta_Y}^{\mathrm{sh}}$ and the local statistics networks $T_{\phi_X}^{\mathrm{sh}}$ and $T_{\phi_Y}^{\mathrm{sh}}$. In constrast to Deep InfoMax \cite{hjelm2018learning}, to enforce to learn only the common information between images $X$ and $Y$, we switch the shared representations to compute the \textit{cross mutual information} as shown in Equation \ref{eq:sharedmutualinformationloss} where global and local mutual information terms are weighted by constant coefficients $\alpha^{\mathrm{sh}}$ and $\beta^{\mathrm{sh}}$. Switching the shared representations is a key element of the proposed method as it enforces to remove the exclusive information of each image (see Section \ref{subsec:Ablation}).
\begin{equation}\label{eq:sharedmutualinformationloss}
\begin{split}
\mathbf{L}_{MI}^{\mathrm{sh}} &= \alpha^{\mathrm{sh}} ( \mathbf{L}^{\mathrm{global}}_{\theta_X, \psi_Y}(X, S_Y) {+} \mathbf{L}^{\mathrm{global}}_{\theta_Y, \psi_X}(Y, S_X) ) \\ 
& \quad + \beta^{\mathrm{sh}} \left( \mathbf{L}^{\mathrm{local}}_{\phi_X, \psi_Y}(X, S_Y) {+} \mathbf{L}^{\mathrm{local}}_{\phi_Y, \psi_X}(Y, S_X) \right)
\end{split}
\end{equation}
Additionally, images $X$ and $Y$ must have identical shared representations, \ie $S_X = S_Y$. A simple solution is to minimize the $L_1$ distance between their shared representations as follows
\begin{equation}\label{eq:sharedloss}
\mathbf{L}_{1} = \mathbb{E}_{p(s_x, s_y)} \left[ \lvert S_X - S_Y \rvert \right]
\end{equation}
The objective function to learn the shared representations is a linear combination of the previous loss terms as can be seen in Equation \ref{eq:SharedFinalLoss}, where $\gamma$ is a constant coefficient.
\begin{equation}
\max_{\{\psi, \theta, \phi \}_{X,Y}} \ \mathcal{L}^{\mathrm{shared}} = \mathbf{L}_{MI}^{\mathrm{sh}} - \gamma \mathbf{L}_{1}
\label{eq:SharedFinalLoss}
\end{equation}

\subsection{Exclusive representation learning}\label{subsec:ExclusiveLearning}

So far, our model is able to extract the shared representations $S_X$ and $S_Y$. Let $E_{\omega_X}^{\mathrm{ex}}:\mathcal{X} \rightarrow \mathcal{E_X}$ and $E_{\omega_Y}^{\mathrm{ex}}:\mathcal{Y} \rightarrow \mathcal{E_Y}$ be the encoder functions to extract the exclusive representations $E_X$ and $E_Y$ from images $X$ and $Y$, respectively. To learn these representations, we estimate and maximize the mutual information between the image $X$ and its corresponding representation $R_X$ which is composed of the shared and exclusive representations \ie $R_X = (S_X, E_X)$. The same operation is performed between the image $Y$ and $R_Y = (S_Y, E_Y)$ as shown in Equation \ref{eq:exclusivemutualinformationloss} where $\alpha^{\mathrm{ex}}$ and $\beta^{\mathrm{ex}}$ are constant coefficients. Mutual information is computed by the global statistics networks $T_{\theta_X}^{\mathrm{ex}}$ and $T_{\theta_Y}^{\mathrm{ex}}$ and the local statistics networks $T_{\phi_X}^{\mathrm{ex}}$ and $T_{\phi_Y}^{\mathrm{ex}}$. Since the shared representation remains constant, we enforce the exclusive representation to include the information which is specific to the image and is not captured by the shared representation.
\begin{equation}\label{eq:exclusivemutualinformationloss}
\begin{split}
\mathbf{L}_{MI}^{\mathrm{ex}} &= \alpha^{\mathrm{ex}} ( \mathbf{L}^{\mathrm{global}}_{\theta_X, \omega_X}(X, R_X) {+} \mathbf{L}^{\mathrm{global}}_{\theta_Y, \omega_Y}(Y, R_Y) ) \\
& \quad + \beta^{\mathrm{ex}} \left( \mathbf{L}^{\mathrm{local}}_{\phi_X, \omega_X}(X, R_X) {+} \mathbf{L}^{\mathrm{local}}_{\phi_Y, \omega_Y}(Y, R_Y) \right)
\end{split}
\end{equation}
On the other hand, the representation $E_X$ must not contain information captured by the representation $S_X$ when maximizing the mutual information between $X$ and $R_X$. Therefore, the mutual information between $E_X$ and $S_X$ must be minimized. While mutual information estimation and maximization via Equation \ref{eq:MutualInformationJSEstimator} works well, using statistics networks fails to converge when performing mutual information estimation and minimization. Therefore, we propose to minimize the mutual information between $S_X$ and $E_X$ via a different implementation of Equation \ref{eq:MutualInformationJSEstimator} using an adversarial objective \cite{Goodfellow2014} as shown in Equation \ref{eq:disentangling_gan_x}. A discriminator $D_{\rho_X}$ defined by a neural network of parameters $\rho_X$ is trained to classify representations drawn from $\mathbb{P}_{S_X E_X}$ as fake samples and representations drawn from $\mathbb{P}_{S_X}\mathbb{P}_{E_X}$ as real samples. Samples from $\mathbb{P}_{S_X E_X}$ are obtained by passing the image $X$ through the encoders $E_{\psi_X}^{\mathrm{sh}}$ and $E_{\omega_X}^{\mathrm{ex}}$ to extract $(S_X, E_X)$. Samples from $\mathbb{P}_{S_X}\mathbb{P}_{E_X}$ are obtained by shuffling the exclusive representations of a batch of samples from $\mathbb{P}_{S_X E_X}$. The encoder function $E_{\omega_X}^{\mathrm{ex}}$ strives to generate exclusive representations $E_X$ that combined with $S_X$ look like drawn from $\mathbb{P}_{S_X}\mathbb{P}_{E_X}$. By minimizing Equation \ref{eq:disentangling_gan_x}, we minimize the Jensen-Shannon divergence $\text{D}_{JS} \left(\mathbb{P}_{S_X E_X} \parallel \mathbb{P}_{S_X}\mathbb{P}_{E_X} \right)$ and thus the mutual information between $E_X$ and $S_X$ is minimized. A similar procedure to generate samples of the product of the marginal distributions from samples of the joint probability distribution is proposed in \cite{brakel2017learning,pmlr-v80-kim18b}. In these models, an adversarial objective is used to make each dimension independent of the remaining dimensions of the representation. Instead, we use an adversarial objective to make the dimensions of the shared part independent of the dimensions of the exclusive part.
\begin{equation}
\begin{split}
\mathbf{L}_{\mathrm{adv}}^{X} &= \mathbb{E}_{p(s_x)p(e_x)} \left[ \log D_{\rho_X}(S_X, E_X) \right] \\
& \quad + \mathbb{E}_{p(s_x, e_x)} \left[ \log\left(1 - D_{\rho_X}(S_X, E_X)\right) \right]
\label{eq:disentangling_gan_x}
\end{split}
\end{equation}
Equation \ref{eq:ExclusiveFinalLoss} shows the objective function to learn the exclusive representation which is a linear combination of the previous loss terms where $\lambda_{\mathrm{adv}}$ is a constant coefficient.
\begin{equation}
\max_{\{\omega, \theta, \phi\}_{X,Y}} \ \ \min_{\{\rho\}_{X,Y}} \ \mathcal{L}^{\mathrm{ex}} = \mathbf{L}_{MI}^{\mathrm{ex}} - \lambda_{\mathrm{adv}} (\mathbf{L}_{\mathrm{adv}}^{X} + \mathbf{L}_{\mathrm{adv}}^{Y})
\label{eq:ExclusiveFinalLoss}
\end{equation}

\subsection{Implementation details}\label{subsec:ImplementationDetails}

Concerning the model architecture, we use DCGAN-like encoders \cite{radford2015unsupervised}, statistics networks similar to those used in Deep InfoMax \cite{hjelm2018learning} and a discriminator defined by a fully-connected network with 3 layers. Every network is trained from scratch using batches of 64 image pairs. We use Adam optimizer with a learning rate value of 0.0001. Concerning the loss coefficients, we use $\alpha^{\mathrm{sh}}=\alpha^{\mathrm{ex}}=0.5$, $\beta^{\mathrm{sh}}=\beta^{\mathrm{ex}}=1.0$, $\gamma=0.1$. The coefficient $\lambda_{\mathrm{adv}}$ is analyzed in Section \ref{subsec:SensitivityAnalysis}. The training algorithm is executed on a NVIDIA Tesla P100. More details about the architecture, hyperparameters and optimizer are provided in the supplementary material section. We also provide our code to train the model and run the experiments.

\section{Experiments}\label{sec:Experiments}

\subsection{Datasets}\label{subsec:Datasets}

\begin{figure}[h!]
\centering
\begin{subfigure}[b]{0.111\textwidth}
\centering
\includegraphics[width=\textwidth]{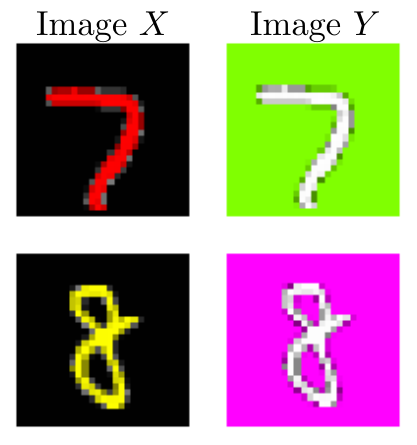}
\caption{}
\label{label0}
\end{subfigure}
\hfill
\begin{subfigure}[b]{0.111\textwidth}
\centering
\includegraphics[width=\textwidth]{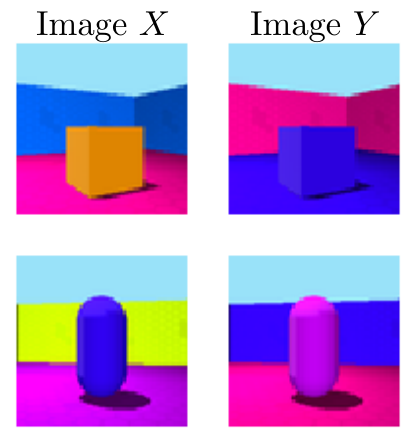}
\caption{}
\label{label1}
\end{subfigure}
\hfill
\begin{subfigure}[b]{0.13\textwidth}
\centering
\includegraphics[width=\textwidth]{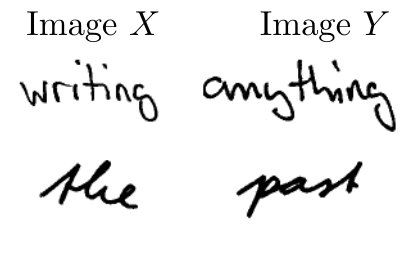}
\caption{}
\label{label2}
\end{subfigure}
\begin{subfigure}[b]{0.111\textwidth}
\centering
\includegraphics[width=\textwidth]{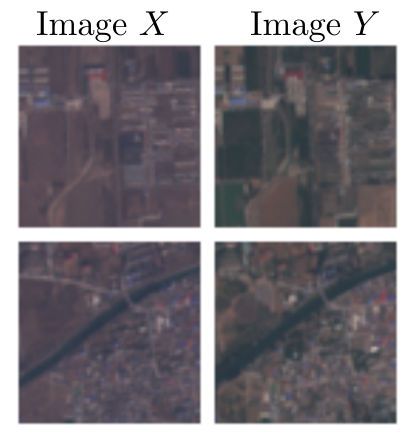}
\caption{}
\label{label1}
\end{subfigure}
\caption{Image pair samples (best viewed in color). (a) Colored-MNIST; (b) 3D Shapes; (c) IAM; (d) Sentinel-2.}
\label{fig:DatasetSamples}
\end{figure}

We perform representation disentanglement on the following datasets: a) \textbf{Colored-MNIST}: Similarly to Gonzalez-Garcia \cite{gonzalez2018image}, we use a colored version of the MNIST dataset \cite{lecun-mnisthandwrittendigit-2010}. The colored background MNIST dataset (MNIST-CB) is generated by modifying the color of the background and the colored digit MNIST dataset (MNIST-CD) is generated by modifying the digit color. The background/digit color is randomly selected from a set of 12 colors. Two images with the same digit are sampled from MNIST-CB and MNIST-CD to create an image pair; b) \textbf{3D Shapes}: The 3D Shapes dataset \cite{3dshapes18} is composed of 480000 images of $64{\times}64{\times}3$ pixels. Each image corresponds to a 3D object in a room with six factors of variation: floor color, wall color, object color, object scale, object shape and scene orientation. These factors of variation have 10, 10, 10, 8, 4 and 15 possible values respectively. We create a new dataset which consists of image pairs where the object scale, object shape and scene orientation are the same for both images while the floor color, wall color and object color are randomly selected; c) \textbf{IAM}: The IAM dataset \cite{marti2002iam} is composed of forms of handwritten English text. Words contained in the forms are isolated and labeled which can be used to train models to perform handwritten text recognition or writer identification. To train our model we select a subset of 6711 images of $64{\times}256{\times}1$ pixels corresponding to the top 50 writers. Our dataset is composed of image pairs where both images correspond to words written by the same person; d) \textbf{Sentinel-2}: Similarly to \cite{SanchezDisentanglingSatellite}, we create a dataset composed of optical images of size $64{\times64}$ from the Sentinel-2 mission \cite{drusch2012sentinel}. A 100GB dataset is created by selecting several regions of interest on the Earth's surface. Image pairs are created by selecting images from the same region but acquired at different times. Further details about the dataset creation can be found on the supplementary material. Some dataset image examples are shown in Figure \ref{fig:DatasetSamples}. For all the datasets, we train our model to learn a shared representation of size 64. An exclusive representation of size 8, 64 and 64 is respectively learned for the colored-MNIST, 3D Shapes and IAM datasets. During training, when data comes from a single domain the number of networks involved can be halved by sharing weights (\ie $\psi_X{=}\psi_Y$, $\theta_X{=}\theta_Y$, etc). For example, the reported results for the 3D Shapes, Sentinel-2 and IAM datasets are obtained using 3 networks (shared representation encoder, global and local statistics networks) to learn the shared representation and 4 networks (discriminator, exclusive representation encoder, global and local statistics networks) to learn the exclusive representation.

\subsection{Representation disentanglement evaluation}\label{subsec:RepresentationEvaluation}

To evaluate the learned representations, we perform several classification experiments. A classifier trained on the shared representation should be good for classifying the shared attributes of the image as the shared representation only contains the common information while it should achieve a performance close to random for classifying the exclusive attributes of the image. An analogous case occurs when performing classification using the exclusive representation. We use a simple architecture composed of 2 hidden fully-connected layers of few neurons to implement the classifier (more details in the supplementary material).

In the colored-MNIST dataset case, a classifier trained on the shared representation must perform well at digit number classification while the accuracy must be close to $8.33\%$ (random decision between 12 colors) at background/digit color classification since no exclusive information is included in the shared representation. Similarly, using the exclusive representations to train a classifier, we expect the classifier to predict correctly the background/digit color while achieving a digit number accuracy close to $10\%$ (random decision between 10 digits) as the exclusive representations contains no digit number information. Results are shown in Tables \ref{tab:MNISTClassificationResultsSharedExclusiveX} and \ref{tab:MNISTClassificationResultsSharedExclusiveY}. We note that the learned representations by our model achieve the expected behavior. The same experiment is performed using the learned representations from the 3D Shapes dataset. A classifier trained on the shared representation must correctly classify the object scale, object shape and scene orientation while the accuracy must be close to random for the floor, wall and object colors ($10\%$, random decision between 10 colors). Differently, a classifier trained on the exclusive representation must correctly classify the floor, wall and object colors while it must achieve a performance close to random to classify the object scale (12.50\%, random decision between 8 scales), object shape (25\%, random decision between 4 shapes) and scene orientation (6.66\%, random decision between 15 orientations). Accuracy results using the shared and exclusive representations are shown in Table \ref{tab:3DShapesClassificationSharedExclusive}.

\begin{table*}
\begin{minipage}{0.52\textwidth}
\centering
  \begin{tabular}{|l|c|c|c|}
    \hline
		\multirow{2}{*}{Method} 								& Background        & Digit       & Distance\\
		      															  	& color             & number      & to ideal\\
    \hline\hline
		Ideal feature $S_X$   				          									& $8.33\%$		                  & $100.00\%$             & $0.0000$\\
		\textbf{Feature $\boldsymbol{S_X}$ (ours)} 								& $\boldsymbol{8.22\%}$		      & $\boldsymbol{94.48\%}$ & $\boldsymbol{0.0563}$\\
		Feature $S_X$ (\cite{gonzalez2018image}) 	                & $99.56\%$ 				            & $95.42\%$ & $0.9581$\\
		\hline\hline
		Ideal feature $E_X$										                    & $100.00\%$ 			              & $10.00\%$              & $0.0000$ \\ 
		\textbf{Feature $\boldsymbol{E_X}$ (ours)}                & $\boldsymbol{99.99\%}$ 			  & $\boldsymbol{13.20\%}$ & $\boldsymbol{0.0321}$\\ 
		Feature $E_X$ (\cite{gonzalez2018image})                  & $99.99\%$ 			              & $71.63\%$              & $0.6164$ \\ 
    \hline
  \end{tabular}
\caption{Background color and digit number accuracy using the shared representation $S_X$ and the exclusive representation $E_X$.}
\label{tab:MNISTClassificationResultsSharedExclusiveX}
\end{minipage}
  ~\hfill~
\begin{minipage}{0.48\textwidth}
\centering
  \begin{tabular}{|l|c|c|c|}
    \hline
		\multirow{2}{*}{Method} 								& Digit             & Digit        & Distance\\
		      															  	& color             & number       & to ideal\\
    \hline\hline
		Ideal feature $S_Y$ 				 										& $8.33\%$  	                  & $100.00\%$ & $0.0000$\\
	  \textbf{Feature $\boldsymbol{S_Y}$ (ours)}  	  & $\boldsymbol{8.83\%}$  	      & $\boldsymbol{94.27\%}$ & $\boldsymbol{0.0623}$\\
    Feature $S_Y$ (\cite{gonzalez2018image}) 	      & $29.81\%$ 	                  & $95.06\%$ & $0.2641$\\
		\hline\hline
		Ideal  feature $E_Y$									           & $100.00\%$               & $10.00\%$  & $0.0000$\\ 
		\textbf{Feature $\boldsymbol{E_Y}$ (ours)}       & $\boldsymbol{99.92\%}$   & $\boldsymbol{13.75\%}$  & $\boldsymbol{0.0383}$\\ 
		Feature $E_Y$ (\cite{gonzalez2018image})         & $99.83\%$                & $74.54\%$  & $0.6471$\\
    \hline
  \end{tabular}
\caption{Digit color and number accuracy using the shared representation $S_Y$ and the exclusive representations $E_Y$.}
\label{tab:MNISTClassificationResultsSharedExclusiveY}
\end{minipage}
\end{table*}

\begin{table*}
\centering
  \begin{tabular}{|l|c|c|c|c|c|c|c|}
    \hline 
		\multirow{2}{*}{Method} 								        & Floor     & Wall      & Object    & Object     & Object     & Scene & Distance\\
		      															  	        & color     & color     & color     & scale      & shape      & Orientation & to ideal \\
    \hline\hline
		Ideal feature $S_X$   				 									& $10.00\%$ & $10.00\%$ & $10.00\%$ & $100.00\%$ & $100.00\%$ & $100.00\%$ & $0.0000$\\
		\textbf{Feature $\boldsymbol{S_X}$ (ours)}      & $\boldsymbol{9.96\%}$  & $\boldsymbol{10.08\%}$ & $\boldsymbol{9.95\%}$  & $\boldsymbol{99.99\%}$  & $\boldsymbol{99.99\%}$  & $\boldsymbol{99.99\%}$ & $\boldsymbol{0.0020}$\\
		Feature $S_X$ (\cite{gonzalez2018image}) 	      & $99.92\%$ & $99.81\%$ & $96.67\%$ & $99.99\%$  & $99.99\%$  & $99.99\%$ & $2.6643$ \\
		\hline\hline
		Ideal feature $E_X$   				 									& $100.00\%$ & $100.00\%$ & $100.00\%$ & $12.50\%$  & $25.00\%$  & $6.66\%$ & $0.0000$\\
		\textbf{Feature $\boldsymbol{E_X}$ (ours)}      & $\boldsymbol{95.10\%}$  & $\boldsymbol{99.79\%}$  & $\boldsymbol{96.17\%}$  & $\boldsymbol{17.25\%}$  & $\boldsymbol{30.73\%}$  & $\boldsymbol{6.79\%}$ & $\boldsymbol{0.1955}$\\
		Feature $E_X$ (\cite{gonzalez2018image}) 	      & $99.99\%$  & $99.99\%$  & $99.94\%$  & $99.06\%$  & $99.98\%$  & $99.81\%$ & $2.5477$\\
    \hline
  \end{tabular}
\caption{Accuracy on the 3D Shapes factors using the disentangled representations $S_X$ and $E_X$.}
\label{tab:3DShapesClassificationSharedExclusive}
\end{table*}

For the colored-MNIST and 3D Shapes datasets, we compare our representations to the representations obtained from the model proposed by Gonzalez-Garcia \etal \cite{gonzalez2018image} using their code. In their model, even if the exclusive factors at image generation are controlled by the exclusive representation, the classification experiment shows that representation disentanglement is not correctly performed as the shared representation contains exclusive information and vice versa. In all the cases, the representations of our model are much closer in terms of accuracy to the ideal disentangled representations than the representations from the model of \cite{gonzalez2018image}. We compute the distance to the ideal representation as the $L_1$ distance between the accuracies on data attributes. As representations obtained from generative models are determined by an objective function defined in the image domain, disentanglement constraints are not explicitly defined in the representation domain. Therefore, representation disentanglement is deficiently achieved in generative models. Moreover, our model is less computationally demanding as it does not require decoder functions to go back into the image domain. Training our model on the colored-MNIST dataset takes 20 min/epoch while the model of \cite{gonzalez2018image} takes 115 min/epoch.

For the IAM dataset, as the shared representation must capture the writer style, it must be useful to perform writer recognition while the exclusive representation must be useful to perform word classification. Accuracy results based on these representations can be seen in Table \ref{tab:IAMClassificationResultsSharedExclusive}. Reasonable results are obtained at writer recognition while less satisfactory results are obtained at word classification as it is a more difficult task. To provide a comparison, we use the latent representation of size 128 learned by a VAE model \cite{Kingma2013} (as the model of \cite{gonzalez2018image} fails to converge) to train a classifier for the mentioned classification tasks. Table \ref{tab:IAMClassificationResultsSharedExclusive} shows that the shared representation outperforms the VAE representation for writer recognition and the exclusive representation achieves a similar performance for word classification.

\begin{table*}[h!]
\begin{minipage}{0.5\textwidth}
\centering
  \begin{tabular}{|l|c|c|}
    \hline
		Method                          								& Writer                & Word \\
    \hline\hline
		Ideal feature $S_X$    				 									& $100.00\%$		                      & $\sim 1.00\%$ \\
		Ideal feature $E_X$															& $\sim 2.00\%$		                    & $100.00\%$ \\
		\textbf{Feature $\boldsymbol{S_X}$ (ours)}      & $\boldsymbol{61.64\%}$		          & $\boldsymbol{9.94\%}$\\
	  \textbf{Feature $\boldsymbol{E_X}$ (ours)}      & $\boldsymbol{10.80\%}$  	          & $\boldsymbol{20.88\%}$  \\
		Feature $f_X$ (\cite{Kingma2013}) 	            & $13.77\%$ 				                  & $20.30\%$\\
    \hline
  \end{tabular}
\caption{Writer and word recognition accuracy.}
\label{tab:IAMClassificationResultsSharedExclusive}
\end{minipage}
  ~\hfill~
\begin{minipage}{0.5\textwidth}
\centering
  \begin{tabular}{|l|c|c|}
    \hline
		Method                          								  & Writer                & Word \\
    \hline\hline
		Feature $S_X$ ($N=1$) 											      & $62.65\%$		          & $15.78\%$\\
		Feature $S_X$ ($N=5$) 											      & $64.06\%$		          & $12.96\%$\\
	  Feature $E_X$ ($N=1$) 											      & $19.68\%$  	          & $19.84\%$\\
	  Feature $E_X$ ($N=5$) 											      & $16.87\%$  	          & $19.69\%$\\
		\hline
  \end{tabular}
\caption{Writer and word recognition accuracy using $N$ nearest neighbors.}
\label{tab:AIMClassificationResultsSharedExclusiveNN}
\end{minipage}
\end{table*}

\begin{figure*}[h!]
\centering
\includegraphics[width=1.00\linewidth]{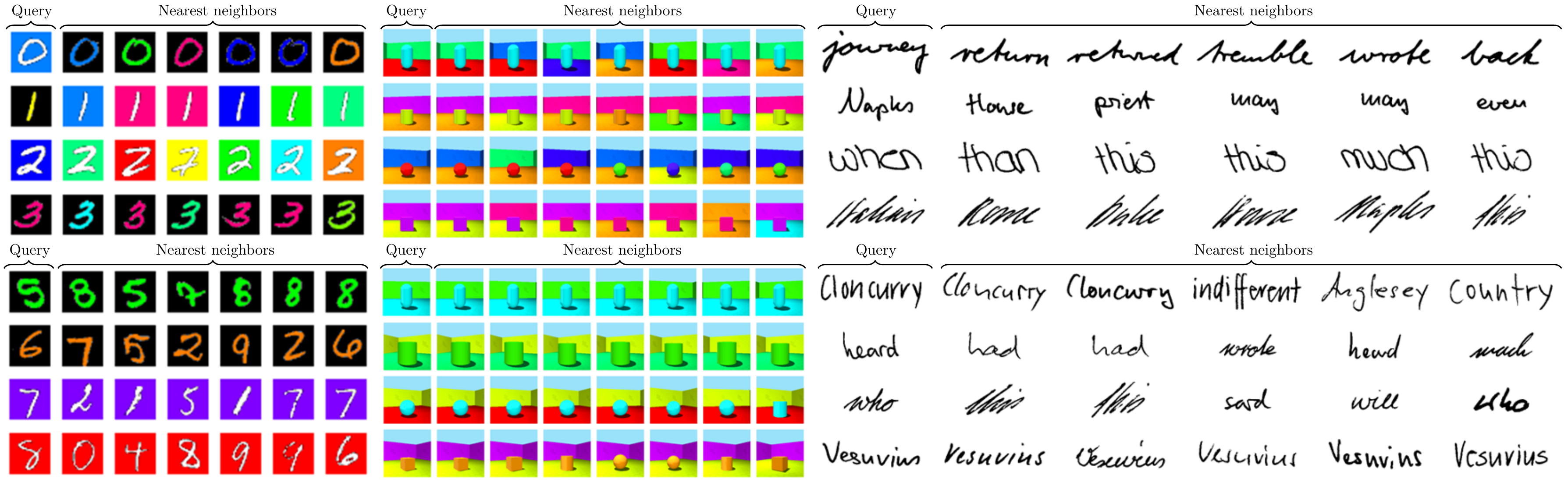}
\caption{Image retrieval on the colored-MNIST, 3D Shapes and IAM datasets (best viewed in color and zoom-in). Retrieved images using the shared representations (on the top) and the exclusive representations (on the bottom).}
\label{fig:ImageRetrievalResultsMNIST3DSHAPES}
\end{figure*}

Additionally, we perform image retrieval experiments using the learned representations. In the colored-MNIST dataset, using the shared representation of a query image retrieves images containing the same digit independently of the background/digit color. In contrast, using the exclusive representation of a query image retrieves images corresponding to the same background/digit color independently of the digit number. A similar case occurs for the 3D Shapes dataset. In the IAM dataset, using the shared representations retrieves words written by the same person or similar style. While using the exclusive representation seems to retrieve images corresponding to the same word. Some image retrieval examples using the shared and exclusive representations are shown in Figure \ref{fig:ImageRetrievalResultsMNIST3DSHAPES}. As image retrieval is useful for clustering attributes, we also perform writer and word recognition on the IAM dataset using $N \in \{1, 5\}$ nearest neighbors based on the disentangled representations. We achieve similar results to those obtained using a neural network classifier as shown in Table \ref{tab:AIMClassificationResultsSharedExclusiveNN}.

\begin{table*}[h!]
\begin{minipage}{0.5\textwidth}
\centering
  \begin{tabular}{|l|c|c|c|}
    \hline
		\multirow{2}{*}{Method} 								& Background        & Digit   & Distance\\
		      															  	& color             & number  & to ideal\\
    \hline\hline
		Ideal feature $S_X$   				 					    & $8.33\%$		                  & $100.00\%$             & $0.0000$\\ 
		\textbf{Baseline}        										& $\boldsymbol{8.22\%}$		      & $\boldsymbol{94.48\%}$ & $\boldsymbol{0.0563}$\\ 
		Baseline (\footnotesize non-SSR)       & $99.99\%$		                  & $89.57\%$              & $1.0209$\\ 
		Baseline ($\gamma=0$)  								    	& $8.49\%$		                  & $92.36\%$              & $0.0780$\\ 
		Baseline ($\alpha^{\mathrm{sh}}=0$) 				& $11.11\%$		                  & $94.83\%$              & $0.0795$\\ 
		Baseline ($\beta^{\mathrm{sh}}=0$) 					& $8.51\%$		                  & $80.59\%$              & $0.1958$\\ 
    \hline
  \end{tabular}
	\caption{MNIST ablation study. Accuracy using $S_X$.}
  \label{tab:MNISTClassificationResultsSharedExclusiveX_ablation}
\end{minipage}
  ~\hfill~
\begin{minipage}{0.5\textwidth}
\centering
  \begin{tabular}{|l|c|c|c|}
    \hline
		\multirow{2}{*}{Method} 								& \multirow{2}{*}{Word}          & \multirow{2}{*}{Writer} & Distance\\
		      															  	&                                &                         & to ideal\\
    \hline\hline
		Ideal feature $S_X$    				 				      & $\sim 1.00\%$              & $100.00\%$                   & $0.0000$\\ 
		\textbf{Baseline} 						      			  & $\boldsymbol{9.94\%}$      & $\boldsymbol{61.64\%}$       & $\boldsymbol{0.4730}$\\ 
		Baseline (\footnotesize non-SSR)       & $20.88\%$                  & $58.94\%$                    & $0.6094$\\ 
		Baseline ($\gamma=0$) 											& $10.51\%$                  & $55.39\%$                    & $0.5412$\\ 
		Baseline ($\alpha^{\mathrm{sh}}=0$) 				& $11.36\%$                  & $61.50\%$                    & $0.4886$\\ 
		Baseline ($\beta^{\mathrm{sh}}=0$) 					& $13.63\%$                  & $50.28\%$                    & $0.6235$\\ 
    \hline
  \end{tabular}
	\caption{IAM ablation study. Accuracy using $S_X$.}
  \label{tab:IAMClassificationResultsSharedExclusive_ablation}
\end{minipage}
\end{table*}

\begin{figure*}[h!]
\centering
\begin{subfigure}[b]{0.32\textwidth}
\centering
\includegraphics[width=\textwidth]{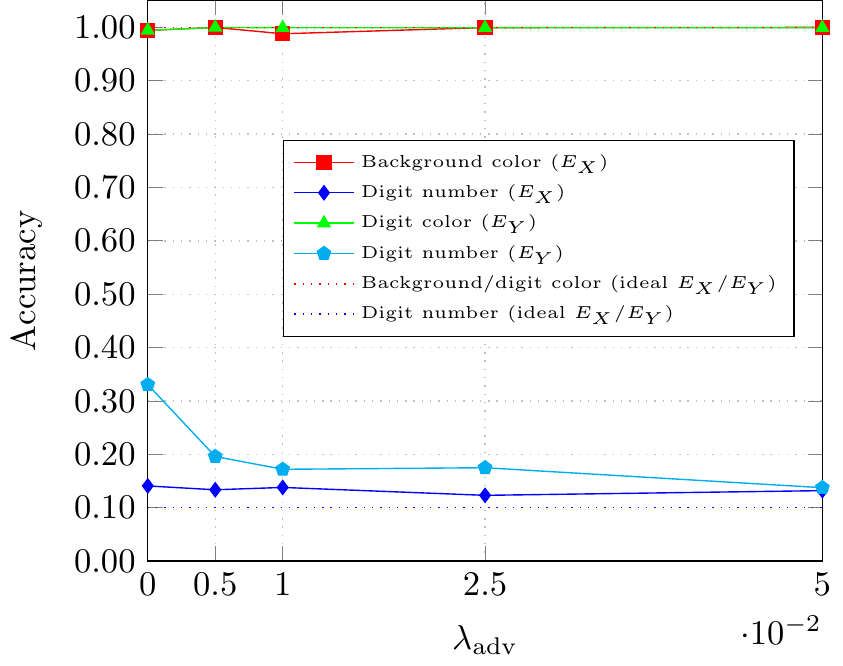}
\caption{}
\label{label0}
\end{subfigure}
\hfill
\begin{subfigure}[b]{0.32\textwidth}
\centering
\includegraphics[width=\textwidth]{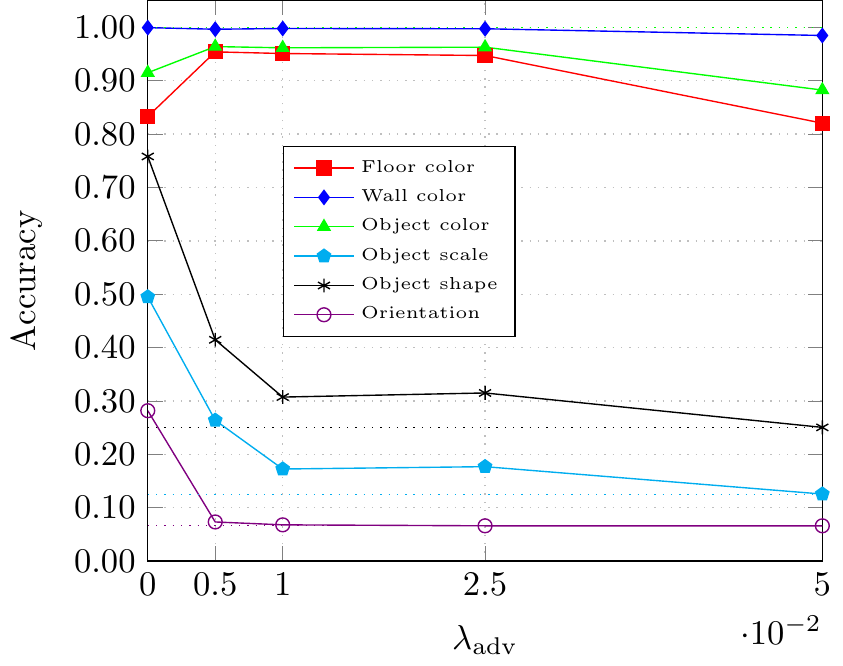}
\caption{}
\label{label1}
\end{subfigure}
\hfill
\begin{subfigure}[b]{0.32\textwidth}
\centering
\includegraphics[width=\textwidth]{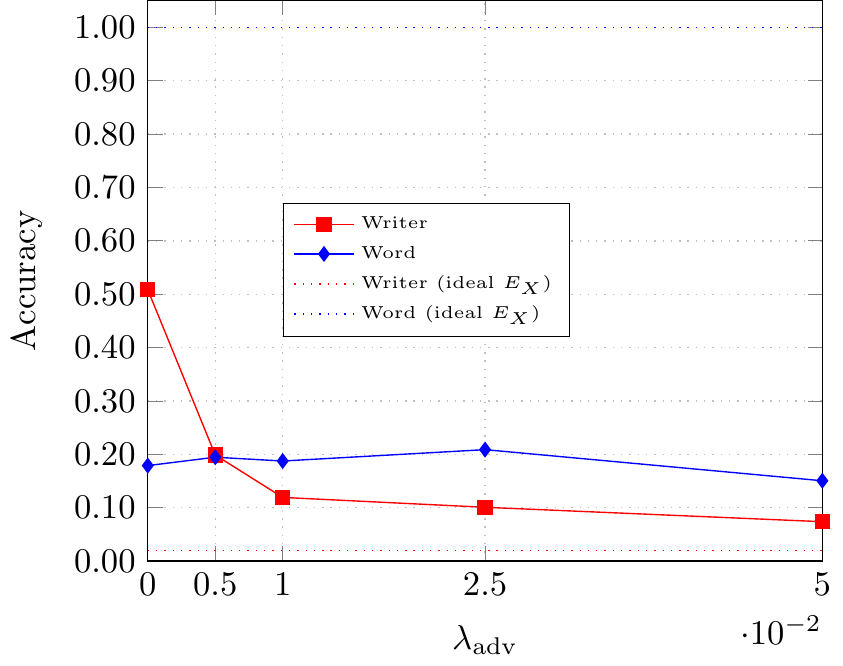}
\caption{}
\label{label2}
\end{subfigure}
\caption{Different values of $\lambda_{\mathrm{adv}}$ are used to learn the exclusive representation. Results are plotted in terms of factor accuracy as a function of $\lambda_{\mathrm{adv}}$. Solid curves correspond to the obtained values and dotted curves correspond to the expected behavior of an ideal exclusive representation (best viewed in color). (a) Colored-MNIST; (b) 3D Shapes; (c) IAM datasets.}
\label{fig:SensitivityAnalysis}
\end{figure*}

\subsection{Analysis of the objective function}
\paragraph{Ablation study}\label{subsec:Ablation}
To evaluate the contribution of each element of the model during the shared representation learning, we remove it and observe the impact on the classification accuracy on the data attributes. As described in Section \ref{subsec:ImplementationDetails}, our baseline setting is the following: $\alpha^{\mathrm{sh}}=0.5$, $\beta^{\mathrm{sh}}=1.0$, $\gamma=0.1$ and switched shared representations $S_X/S_Y$ (SSR). We perform the ablation study and show the results for the colored-MNIST and IAM datasets in Tables \ref{tab:MNISTClassificationResultsSharedExclusiveX_ablation} and \ref{tab:IAMClassificationResultsSharedExclusive_ablation}. Switching the shared representations plays a crucial role in representation disentanglement avoiding these representations to capture exclusive information. When the shared representations are not switched (non-SSR), the accuracy on exclusive attributes considerably increases meaning the presence of exclusive information in the shared representations. Removing the $L_1$ distance between $S_X$ and $S_Y$ ($\gamma=0$) slightly reduces the accuracy on shared attributes. Removing the global mutual information term ($\alpha^{\mathrm{sh}}=0$) slightly increases the presence of exclusive information in the shared representation. Finally, using the local mutual information term is important to capture the shared information as the accuracy on shared attributes considerably decreases when setting $\beta^{\mathrm{sh}}=0$. Similar results are obtained by setting $\alpha^{\mathrm{ex}}=0$ or $\beta^{\mathrm{ex}}=0$ during the exclusive representation learning. In general, all loss terms lead to an improvement in representation disentanglement.
\paragraph{Sensitivity analysis}\label{subsec:SensitivityAnalysis}
As the parameter $\lambda_{\mathrm{adv}}$ weights the term that minimizes the mutual information between the shared and exclusive representations, we empirically investigate the impact of this parameter on the information captured by the exclusive representation. We use different values of $\lambda_{\mathrm{adv}} \in \{0.0, 0.005, 0.010, 0.025, 0.05\}$ to train our model. Then, exclusive representations are used to perform classification on the attributes of data. Results in terms of accuracy as a function of $\lambda_{\mathrm{adv}}$ are shown in Figure \ref{fig:SensitivityAnalysis}. For $\lambda_{\mathrm{adv}}=0.0$ no representation disentanglement is performed, then the exclusive representation contains shared information and achieves a classification performance higher than random for the shared attributes of data. While increasing the value of $\lambda_{\mathrm{adv}}$ the exclusive representation behavior (solid curves) converges to the expected behavior (dotted curves). However, values higher than $0.025$ decrease the performance classification on exclusive attributes of data.

\subsection{Satellite applications}

We show that our model is particularly useful when large amounts of unlabeled data are available and labels are scarce as in the case of satellite data. We train our model to learn the shared representations of our Sentinel-2 dataset which contains 100GB of unlabeled data. Then, a classifier is trained on the EuroSAT dataset \cite{DBLP:journals/corr/abs-1709-00029} (27000 Sentinel-2 images of size $64{\times64}$ labeled in 10 classes) using the learned representations of our model as inputs. While training a classifier on the shared representation, we make it robust to time-related conditions (seasonal changes, atmospheric conditions, etc.). We achieve an accuracy of $93.11\%$ outperforming the performance obtained using the representations of a VAE model \cite{Kingma2013} ($87.64\%$) and the VAE-GAN model proposed by Sanchez \etal \cite{SanchezDisentanglingSatellite} ($92.38\%$).

\begin{figure}[h!]
\centering
\includegraphics[width=1.00\linewidth]{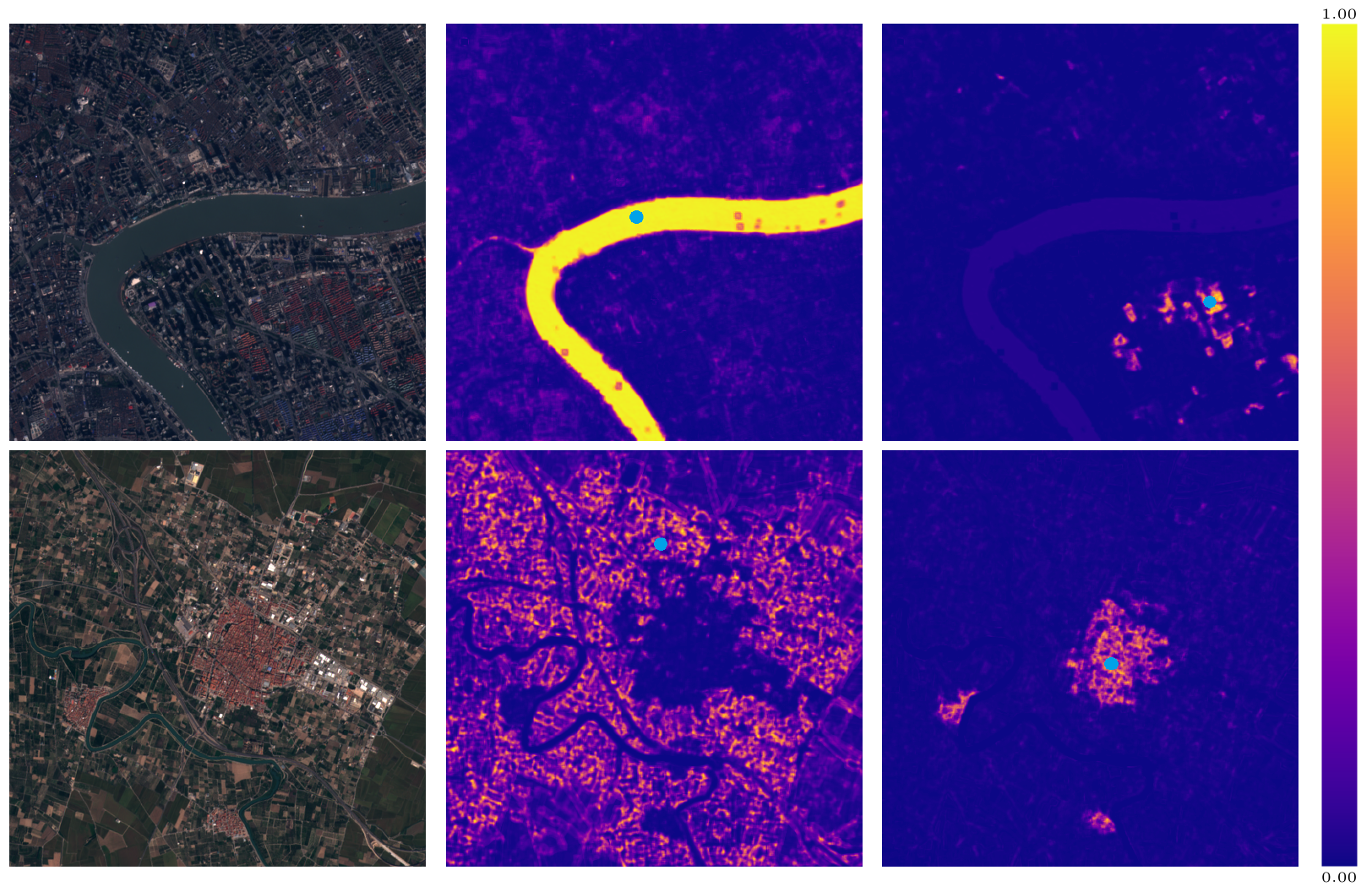}
\caption{Pixel distance based on mutual information. The mutual information is computed between a given pixel (blue point) and the remaining image pixels via Equation \ref{eq:sharedmutualinformationloss}.}
\label{fig:ImageRetrievalResultsMNIST3DSHAPESentinel}
\end{figure}

As another interesting application, we found that Equation \ref{eq:sharedmutualinformationloss} could be used to measure the distance between the center pixels of image patches $X$ and $Y$ in terms of mutual information. Some examples are shown in Figure \ref{fig:ImageRetrievalResultsMNIST3DSHAPESentinel}. As can be seen, using this distance we are able to distinguish the river, urban regions and agricultural areas. We think this could be useful for further applications such as unsupervised image segmentation and object detection.

\section{Conclusions}\label{sec:Conclusions}
We have proposed a novel method to perform representation disentanglement on paired images based on mutual information estimation using a two-stage training procedure. We have shown that our model is less computationally demanding and outperforms the VAE-GAN model of \cite{gonzalez2018image} to disentangle representations via classification experiments in three datasets. Additionally, we have performed an ablation study to demonstrated the usefulness of the key elements of our model (switched shared representations, local and global statistics networks) and their impact on representation disentanglement. Then, we have empirically proven the disentangling capability of our model by analyzing the role of $\lambda_{\mathrm{adv}}$ during training. Finally, we have demonstrated the benefits of our model on a challenging setting where large amounts of unlabeled paired data are available as in the Sentinel-2 case. We have shown that our model outperforms models relying on image reconstruction or image generation at image classification. We have also shown that the \textit{cross mutual information} objective could be useful for unsupervised image segmentation and object detection.

{\small
\bibliographystyle{ieee_fullname}
\bibliography{egbib}

\begin{thebibliography}{10}\itemsep=-1pt

\bibitem{arjovsky2017wasserstein}
Martin Arjovsky, Soumith Chintala, and L{\'e}on Bottou.
\newblock {W}asserstein generative adversarial networks.
\newblock In {\em Proceedings of the 34th International Conference on Machine
  Learning}, 2017.

\bibitem{belghazi2018mine}
Mohamed~Ishmael Belghazi, Aristide Baratin, Sai Rajeshwar, Sherjil Ozair,
  Yoshua Bengio, Aaron Courville, and Devon Hjelm.
\newblock Mutual information neural estimation.
\newblock In {\em Proceedings of the 35th International Conference on Machine
  Learning}, 2018.

\bibitem{brakel2017learning}
Philemon Brakel and Yoshua Bengio.
\newblock Learning independent features with adversarial nets for non-linear
  ica.
\newblock {\em arXiv preprint arXiv:1710.05050}, 2017.

\bibitem{3dshapes18}
Chris Burgess and Hyunjik Kim.
\newblock 3d shapes dataset.
\newblock https://github.com/deepmind/3dshapes-dataset/, 2018.

\bibitem{Chen2016}
Xi Chen, Yan Duan, Rein Houthooft, John Schulman, Ilya Sutskever, and Pieter
  Abbeel.
\newblock Infogan: Interpretable representation learning by information
  maximizing generative adversarial nets.
\newblock In {\em Advances in Neural Information Processing Systems}, 2016.

\bibitem{donahue2016adversarial}
Jeff Donahue, Philipp Kr{\"a}henb{\"u}hl, and Trevor Darrell.
\newblock Adversarial feature learning.
\newblock In {\em International Conference on Learning Representations}, 2017.

\bibitem{drusch2012sentinel}
Matthias Drusch, Umberto Del~Bello, S{\'e}bastien Carlier, Olivier Colin,
  Veronica Fernandez, Ferran Gascon, Bianca Hoersch, Claudia Isola, Paolo
  Laberinti, Philippe Martimort, et~al.
\newblock Sentinel-2: Esa's optical high-resolution mission for gmes
  operational services.
\newblock {\em Remote sensing of Environment}, 120:25--36, 2012.

\bibitem{ganin2014unsupervised}
Yaroslav Ganin and Victor Lempitsky.
\newblock Unsupervised domain adaptation by backpropagation.
\newblock In {\em Proceedings of the 32nd International Conference on Machine
  Learning}, 2015.

\bibitem{gonzalez2018image}
Abel Gonzalez-Garcia, Joost van~de Weijer, and Yoshua Bengio.
\newblock Image-to-image translation for cross-domain disentanglement.
\newblock In {\em Advances in Neural Information Processing Systems}. 2018.

\bibitem{Goodfellow2014}
Ian Goodfellow, Jean Pouget-Abadie, Mehdi Mirza, Bing Xu, David Warde-Farley,
  Sherjil Ozair, Aaron Courville, and Yoshua Bengio.
\newblock Generative adversarial nets.
\newblock In {\em Advances in neural information processing systems}, 2014.

\bibitem{Goodfellow2016}
Ian~J. Goodfellow.
\newblock {NIPS} 2016 tutorial: Generative adversarial networks.
\newblock 2016.

\bibitem{DBLP:journals/corr/abs-1709-00029}
Patrick Helber, Benjamin Bischke, Andreas Dengel, and Damian Borth.
\newblock Eurosat: {A} novel dataset and deep learning benchmark for land use
  and land cover classification.
\newblock {\em CoRR}, abs/1709.00029, 2017.

\bibitem{higgins2017beta}
Irina Higgins, Loic Matthey, Arka Pal, Christopher Burgess, Xavier Glorot,
  Matthew Botvinick, Shakir Mohamed, and Alexander Lerchner.
\newblock beta-vae: Learning basic visual concepts with a constrained
  variational framework.
\newblock In {\em International Conference on Learning Representations}, 2017.

\bibitem{hjelm2018learning}
R~Devon Hjelm, Alex Fedorov, Samuel Lavoie-Marchildon, Karan Grewal, Phil
  Bachman, Adam Trischler, and Yoshua Bengio.
\newblock Learning deep representations by mutual information estimation and
  maximization.
\newblock In {\em International Conference on Learning Representations}, 2019.

\bibitem{karras2018style}
Tero Karras, Samuli Laine, and Timo Aila.
\newblock A style-based generator architecture for generative adversarial
  networks.
\newblock In {\em Proceedings of the IEEE Conference on Computer Vision and
  Pattern Recognition}, 2019.

\bibitem{pmlr-v80-kim18b}
Hyunjik Kim and Andriy Mnih.
\newblock Disentangling by factorising.
\newblock In {\em Proceedings of the 35th International Conference on Machine
  Learning}, 2018.

\bibitem{Kingma2013}
Diederik~P Kingma and Max Welling.
\newblock Auto-encoding variational bayes.
\newblock In {\em International Conference on Learning Representations}, 2014.

\bibitem{pmlr-v48-larsen16}
Anders Boesen~Lindbo Larsen, Søren~Kaae Sønderby, Hugo Larochelle, and Ole
  Winther.
\newblock Autoencoding beyond pixels using a learned similarity metric.
\newblock In {\em Proceedings of The 33rd International Conference on Machine
  Learning}, 2016.

\bibitem{lecun-mnisthandwrittendigit-2010}
Yann LeCun and Corinna Cortes.
\newblock {MNIST} handwritten digit database.
\newblock 2010.

\bibitem{Mao2017}
Xudong Mao, Qing Li, Haoran Xie, Raymond~YK Lau, Zhen Wang, and Stephen~Paul
  Smolley.
\newblock Least squares generative adversarial networks.
\newblock In {\em 2017 IEEE International Conference on Computer Vision
  (ICCV)}, 2017.

\bibitem{marti2002iam}
U-V Marti and Horst Bunke.
\newblock The {IAM}-database: an english sentence database for offline
  handwriting recognition.
\newblock {\em International Journal on Document Analysis and Recognition},
  2002.

\bibitem{oord2018representation}
Aaron van~den Oord, Yazhe Li, and Oriol Vinyals.
\newblock Representation learning with contrastive predictive coding.
\newblock {\em arXiv preprint arXiv:1807.03748}, 2018.

\bibitem{ozair2019wasserstein}
Sherjil Ozair, Corey Lynch, Yoshua Bengio, Aaron van~den Oord, Sergey Levine,
  and Pierre Sermanet.
\newblock Wasserstein dependency measure for representation learning.
\newblock {\em arXiv preprint arXiv:1903.11780}, 2019.

\bibitem{park2019SPADE}
Taesung Park, Ming-Yu Liu, Ting-Chun Wang, and Jun-Yan Zhu.
\newblock Semantic image synthesis with spatially-adaptive normalization.
\newblock In {\em Proceedings of the IEEE Conference on Computer Vision and
  Pattern Recognition}, 2019.

\bibitem{radford2015unsupervised}
Alec Radford, Luke Metz, and Soumith Chintala.
\newblock Unsupervised representation learning with deep convolutional
  generative adversarial networks.
\newblock In {\em International Conference on Learning Representations}, 2016.

\bibitem{SanchezDisentanglingSatellite}
Eduardo Sanchez, Mathieu Serrurier, and Mathias Ortner.
\newblock Learning disentangled representations of satellite image time series.
\newblock In {\em Proceedings of the European Conference on Machine Learning
  and Principles and Practice of Knowledge Discovery in Databases}, 2019.

\bibitem{tran2017disentangled}
Luan Tran, Xi Yin, and Xiaoming Liu.
\newblock Disentangled representation learning gan for pose-invariant face
  recognition.
\newblock In {\em Proceedings of the IEEE Conference on Computer Vision and
  Pattern Recognition}, 2017.

\end{thebibliography}
}

\end{document}